\documentclass[letterpaper]{article}
\usepackage{aaai}
\usepackage{times}
\usepackage{helvet}
\usepackage{courier}

\usepackage[utf8]{inputenc} % allow utf-8 input
\usepackage[T1]{fontenc}    % use 8-bit T1 fonts
\usepackage{lineno, hyperref}       % hyperlinks
\usepackage{url}            % simple URL typesetting
\usepackage{booktabs}       % professional-quality tables
\usepackage{amsfonts}       % blackboard math symbols
\usepackage{nicefrac}       % compact symbols for 1/2, etc.
\usepackage{microtype}      % microtypography
\usepackage{amsmath, amssymb, amsthm, amsfonts}
\usepackage{multirow}
\usepackage{threeparttable}
\usepackage{graphicx}
\usepackage{caption}
\usepackage{subfigure}
\usepackage{wrapfig}
\usepackage{color}
\begin{document}

\title{Metabolize Neural Network}
\author{Dan Dai, Zhiwen Yu, Yang Hu, Wenming Cao, Mingnan Luo\\
School of Computer Science \& Engineering, South China University of Technology\\
\{csdaidan@mail., zhwyu@., cssuperhy@mail., csluomingnan@mail.\}scut.edu.cn, wenmincao2-c@my.cityu.edu.hk
}

\maketitle
\begin{abstract}
\begin{quote}
The metabolism of cells is the most basic and important part of human function. Neural networks in deep learning stem from neuronal activity. It is self-evident that the significance of metabolize neuronal network(MetaNet) in model construction. In this study, we explore neuronal metabolism for shallow network from proliferation and autophagy two aspects. First, we propose different neuron proliferate methods that constructive the self-growing network in metabolism cycle. Proliferate neurons alleviate resources wasting and insufficient model learning problem when network initializes more or less parameters. Then combined with autophagy mechanism in the process of model self construction to ablate under-expressed neurons. The MetaNet can automatically determine the number of neurons during training, further, save more resource consumption. We verify the performance of the proposed methods on datasets: MNIST, Fashion-MNIST and CIFAR-10.
\end{quote}
\end{abstract}

\section{Introduction}

\noindent Human cells and tissues by proliferation and autophagy to build freah ones and replacing outdated, living organisms are needed to support metabolism. Cell proliferation and autophagy are the mechanisms to grow and degrade structures, it can be a self-proliferation, protection or procedural death mechanism ~\cite{heiden2009understanding,rabinowitz2010autophagy}. The theory development of neural networks originates from biology, the activity of neuronal in deep network should analogy with the metabolism in cells. The neurons activity foundations in the model guarantee the vitality of generalization.

Deep neural networks with some hidden layers have led to the substantial tremendous progress in many applications. The deep networks are trained with initialize various parammeters, like weights, bias, the neurons number in each layer, etc. Those networks suffer from the waste of computing resources when initialize a large number of model parameters and construct complex structure in training process. However, the network function limited when the data mapping simply by initialize few parameters can not be learned with efficiently results. Networks which can depict different data structures is extremely important, it are easily overfit the data when networks are large, lack of learning ability when simply construct the network structures.

Ideally, what is desired is a network large enough to learn the function mapping, and as small as possible to generalize well ~\cite{huyser1988generalization}. There are destructive and constructive two general approaches to finding such networks. The destructive is using a larger and deeper than needed network, training it until meet the required tasks. After this, elements of the network are pruned off if they have little influence on result, The constructive approach starts with a small network and grows until a solution is found ~\cite{ash1989dynamic,alvarez2016learning}.

In this work we study techniques for how to metabolize the network with proliferate new and phagocytose decayed neurons by exploiting the fact that the network how to self-regulation. The method we present is a dynamic self-growth network with general applicablity. Our MetaNet take advantage of the growth network to scientifically and comprehensively deduce the development behavior of shallow neural network as we learn from it. Given a small and initial model, we first proliferate some new hidden neurons that using static or dynamic technique, similar to the meta learning methods ~\cite{sung2017learning,sohn2012learning}, then utilize autoregression technique to set weights for those introducted neurons. Further, consider the combining pre-growth network to phagocytose decayed neurons based on various metrics.

The contributions of this paper are threefold. First, We design the ideology of MetaNet based on the cell metabolic mechanism, and explore neurons metabolism for shallow network from proliferation and autophagy two aspects. Second, we propose various methods of proliferate and phagocytose neurons in the metabolism process, which are not affect the training process of the overall network as far as possible. Finally, we make some experiments to verify our ideas on MNIST, Fashion MNIST and CIFAR-10 datasets, which compare and analyse the results of diverse proliferation and autophagy methods.

\section{Related Work}

\noindent Model growing for neural network, such as deep growing learning, network pruning, network distillation and weights learning have been explored in many literature, but the neurous number in each layer, and how to proliferate additional and phagocytose failed neurous starts with a small network has not been widely studied. Currently, most works about this part is achieved by manually tuning hyper-parameters, or by relying on a trained network that cost resources. Furthermore, if the amount of initial hyper-parameter are small, the network lack of adequate learning ability, conversely, most of the parameters are redundant and resource consuming ~\cite{cheng2015an}.

The evolution of network structure have destructive and constructive two general approaches, destructive methods to structure evolution be required for a pretrained large and deep network, then prune or mimic it to network self-repairing or forming a new shallower or thinner student network. Network pruning by~\cite{hanson1989comparing,lecun1990optimal,whitley1990genetic,reed1993pruning,setiono1997neural} delete redundant parameters to improve network generalization ability. Recent explored ~\cite{han2016deep} have introduced deep compression by pruning, trained quantization and Huffman coding. Anwar S \textit{et al}.~\cite{anwar2017structured} proposed structured pruning in deep convolutional neural networks(DCNN) which made channel wise, kernel wise and intra kernel strided structured sparsity,  ~\cite{anwar2017coarse} showed layer-wise pruning, feature map pruning, $k \times k$ kernel pruning and intra-kernel pruning four possible pruning granularities, used pruning to reduce the computational complexity of a DCNN. Optimizing the number of neurons in a network ~\cite{liu2007optimizing,murray2015auto-sizing,zhou2016less,alvarez2016learning}. The core concept of these studies are remove some network elements that have less effect on the model performance.

The main idea of model compression ~\cite{bucilua2006model} is used a fast and compact model to approximate the function learned by a slower, larger, but better performing model. This concept has been extended in ~\cite{hinton2015distilling} to formed network distillation, which transfered the knowledge from the cumbersome model to a small model that is more suitable for deployment. The more recent works often to training shallower or thinner model when distilled the acquired knowledge from a large model. ~\cite{ba2014do} desinged a shallow feed-forward nets that could learned the complex functions from trained deep nets. ~\cite{chen2016net2net} and ~\cite{li2017learning} learned a continually growing model by pretrained network knowledge.

Another constructive method is dynamic growthing neurons or layers for initial small network structure that satisfies certain conditions. For add new neurons or layers,  ~\cite{ash1989dynamic} showed dynamic node creation that automatically grows backpropagation networks. ~\cite{rusu2016progressive} by combining transfer analysis and reinforcement learning, progressive network achieved old and new features to multiple tasks.
~\cite{terekhov2015knowledge} added an additional block when learned to solve new tasks. These network expansion by learning between different tasks. Not only are these processes increasing nodes, but need to consider the setting of the parameters. As Glorot and Bengion ~\cite{glorot2010understanding} and ~\cite{he2015delving} proposed a new initialization scheme for different activation function, respectively. ~\cite{denil2013predicting} predicted parameters in deep network. ~\cite{wang2017deep} proposed deep growing learning framework by selected the confident prediction examples as the next iteration.

The destructive methods have higher requirements for prior knowledge or pretrained model, take major resources in trained a network that is larger than necessary, it may get
stuck in one of the intermediately sized optimal solutions ~\cite{ash1989dynamic}. To construction, on the existing conditions to growth when add additional neurons or layers. In this paper, we introduce metabolism ideology into neural network, use proliferation and autophagy to gradually self-regulation in different model training stages.

\section{The Proposed Method}

\begin{figure}
\centerline{{\includegraphics[scale=0.6,clip,keepaspectratio]{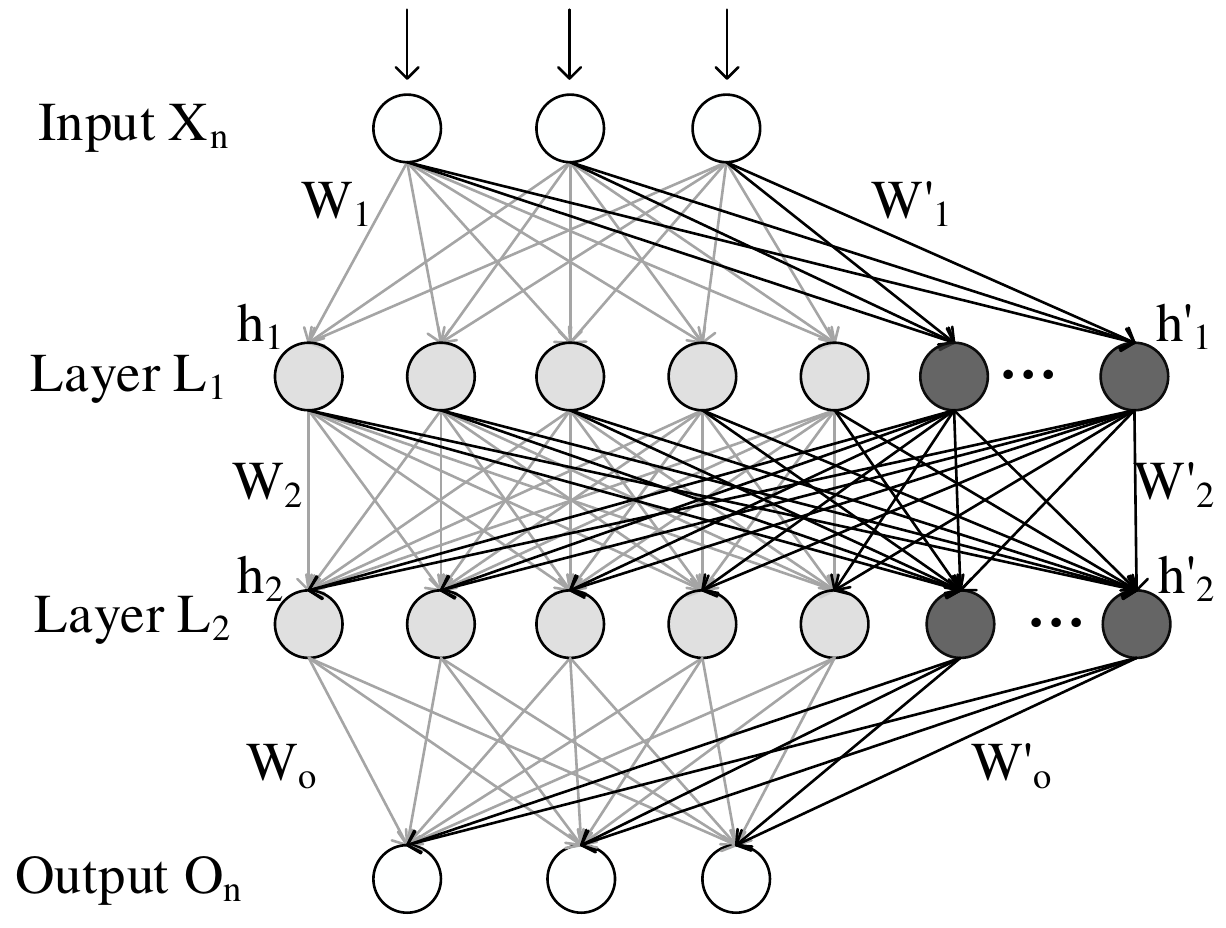}}}
\caption{Illustration of a dynamic development feedforward network with two hidden layers. The hollow units represented input and output, light colour solid units are original neurons, deep colour solid units are the new neurons which dynamic increasing from original neurons. $W$ and $h$ are weights and output of each layer, respectively.}
\label{fig_example1.ps}
\end{figure}

\begin{figure*}
\centerline{{\includegraphics[scale=0.6,clip,keepaspectratio]{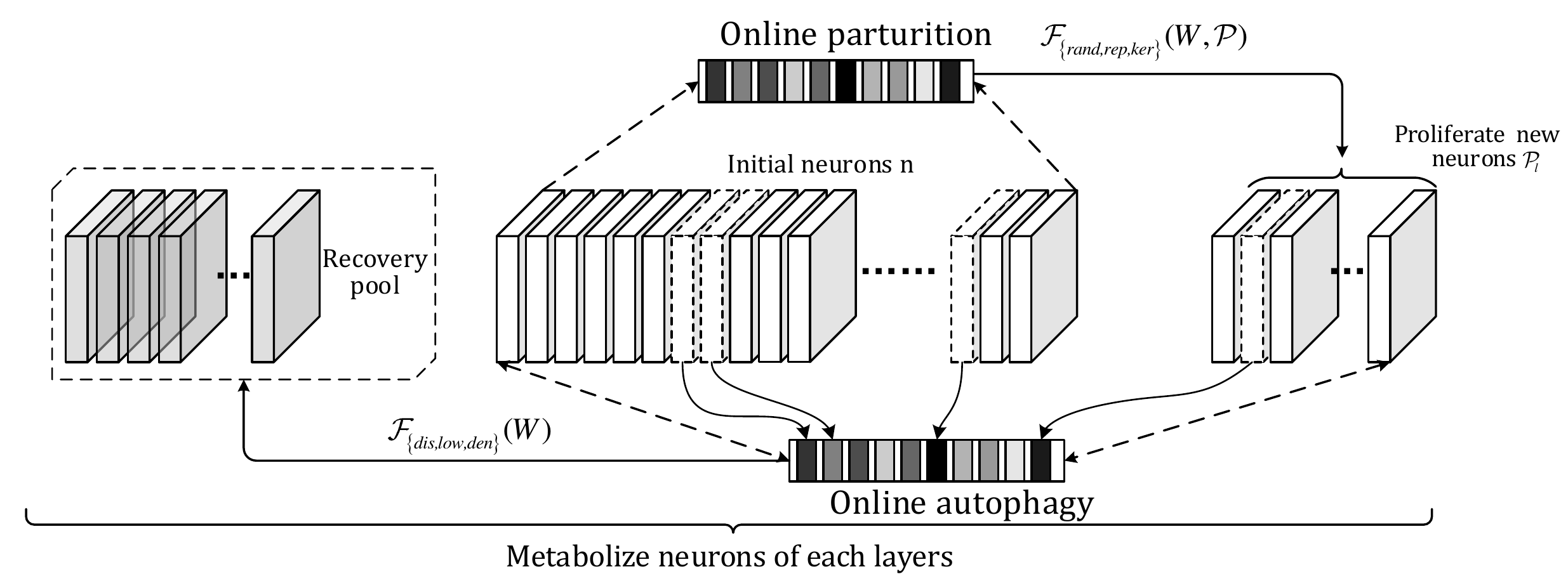}}}
\caption{Illustration of how to proliferate and phagein neurons in each hidden layers}
\label{fig_example2.ps}
\end{figure*}

Our model only need relatively small retraning effort when some new neurons is introduced, the overall framework is show in Figure\ref{fig_example1.ps}, and detailed diagram in Figure \ref{fig_example2.ps}. This section we first introduce metabolism cycle, then proliferate neurons number in each layer, descibe weight learning for adding neurons and discuss the autophagy process in MetaNet.

Given a simple initialize network, the main parameters in the model are expressed as $n=\{n_1,n_2,\cdots,n_l\}$, $W=\{W_1,W_2,\cdots,W_l\}$, $h=\{h_{1},h_{2},\cdots,h_{l} \} $, neurons number $n_l$ , weights $W_l$ and each hidden output $h_l$ in layer $L_l$.

\subsection{Metabolism Cycle}

In MetaNet, some new hidden neurons are introducted when the curve slope of loss function value $\mathcal{L}$ in metabolism cycle $\Delta c = c_{g}-c_{b}$ begins to lower than $\varpi$, it means that the initialization weights couldn't satisfied network learning ability, where $ c_{g} $ and $ c_{b} $ are the epoch in training process, the mathematically expressed as:

\begin{equation}
\frac{\mathcal L(c_{b}) - \mathcal L(c_{g}) }{\Delta c} \leq \varpi
\end{equation}

At first, the initialization model need longer cycle to training, and maintain the sufficient training for new neurons, the $\Delta c$ shouldn't be too small, $c_{g}$ and $c_{b}$ will continuous change with epochs time, the metabolism cycle $\Delta c$ also can changes depending on the situation.

\subsection{Proliferate Neurons}
The proliferate neurons number $ \mathcal P=\{\mathcal P_1,\mathcal P_2,\cdots,\mathcal P_l\}$, introducted $P_l$ new nodes in each layer $L_l$ . There should be  two basic approaches to decide adding neurons numbers, one is static methods that using a constant value to network grow and training it until end the iteration process, an alternative approach is dynamically increasing that as some variables grow automatically, in this section, we select the variance of test error $\mathit{Var}(E)$ in $\Delta c$ cycle period as the independent variable of the number of neurons in all layers, where $E=\{E_b,\cdots,E_g\}$, $E_n$ is the test error at $epoch\ n$. The proliferation neurons operation as $\mathit{ProN}$, the number of neurons is given by:
\begin{equation}
\mathcal P_l = c_l(1 + e^{ \mathit f (\mathit{Var}(E),\vartheta_l)})
\end{equation}
Here $ c_l $ is the constant value in each layer $L_l$ and $ \vartheta_l $ is the dynamic proliferation rate. This function $ \mathit f $ can static or dynamic to obtain the neurons number $\mathcal P_l$ rely on the linear or nonlinear relation between $\mathit{Var}(E)$ and parameter $ \vartheta_l $. In our model, we make $ \mathit f (\mathit{Var}(E),\vartheta_l) =-\vartheta_l \mathit{Var}(E)$, where $ \vartheta_l=1 $.

\subsection{Weights Learning}

Training network with a large number of neurons and parameters from beginning to end, each neurons may not try to play their due role and function. While initialize a large number of neurons at the first, would bring a little free parameters number, exploiting the weights structure of each layers in deep network is necessary. To such problems is to define a autoregression model for the prior weights, the autoregression weights function $\mathcal G_l:{ \mathbb R^ {n_l} \mapsto \mathbb{R} }$, generative neuron weights as $W_l^{'}= \mathcal G_l(W_l,\mathcal P_l)$, $W^{'}=\{W_1^{'},W_2^{'},\cdots,W_l^{'}\}$. In this section, we will introduce three methods of weight learning.

\textbf{Random Parameterization}
For the weights obtained from the initial rounds of training, function $\mathcal{F}_{rand}$ is a randomly select the number of $\mathcal P_l$ neuron weights from original weights $W$ as newly generated neuron weights $W^{'}=\{W_1^{'},W_2^{'},\cdots,W_{P_l}^{'},\}$, this process is expressed as:

\begin{equation}
W^{'}=\mathcal{F}_{rand} (W,\mathcal P)
\end{equation}

\textbf{Higher Representational Power of Neurons}
For training samples, the ability of each neuron to express may reflect its value to some extent. Each single neurons $N_s $ has an activations values $ \varphi (N_s)=\{\varphi_1,\varphi_2,\cdots,\varphi_n \} $ and representation power $\mathcal R (N_s,X) =\mathit{Var}(\varphi (N_s))$ for different input $X=\{X_1,X_2,\cdots,X_n\}$. Select the top $\mathcal P_l$ with the highest variance values from $\mathcal R_l (N_s,X) $ in layer $l$ to form new weights.

\begin{equation}
W^{'}=\mathcal{F}_{rep} (\mathcal R (N_s,X),\mathcal P)
\end{equation}

\textbf{Autoregression Strategy}
In the process of proliferate neurons, our model start with a small amount of original neuron weights to get new weights. These weights as initial samples, we can use meta-learning methods to obtain some new samples. ~\cite{denil2013predicting} analysed the low rank weight meatics and proposed three ways to constructed an approprite dictionary: trained a single layer unsupervised model, exploit prior knowledge about the structure of features space and used kernel to encode prior knowledge. On the basis of this study, we use kernel ridge regressive\cite{Murphy2012MachineLearning} to make online weights prediction, as it is a autoregressive new parameter in all weight regions. This method is expressed as:

\begin{equation}
W^{'}=\mathcal{F}_{ker}(K^T(K+\lambda I)^{-1}W,\mathcal P)
\end{equation}

\begin{equation}
K(w_i,w_j)=exp(-\frac{{\|w_i-w_j\|}^2}{2\delta ^2})
\end{equation}
Here kernel matrix $K_n$,  $\delta$ is known as the bandwidth that controls the degree of smoothness.

\subsection{Neurons Selection for Autophagy}
Survival of the fittest in the neurons dynamic growth is the effective mode of network development. The question is just how many neurons are needed for a particular problem in design the dynamic growth network, what conditions does these neurons meet and whether they have any help in training process. In this section, we propose three methods of neurons autophagy.

\textbf{Weight Distribution}
Weight plays a key role in model training. As Glorot and Bengion ~\cite{glorot2010understanding} found that the weights are initialized independently and the inputs features variances are the same with $\mathit{Var}(X)$. For the function Relu and forward propagation, with $L$ layers put together, the weight variance of each layer should satisfy the following conditions ~\cite{he2015delving}:
\begin{equation}
Var(h_l)=Var(h_1)(\prod\limits_{l = 1}^{L}\frac{1}{2}n_l\mathit{Var}(W_l))
\end{equation}

\begin{equation}
\frac{1}{2}n_l\mathit{Var}(W_l)=1
\end{equation}
It is constantly in a condition of growthing and updating. If the weight is small, the effect may not be obvious, use distribution $\mathcal T(\beta) $ as absolute value less than threshold $\beta$. Before neurons proliferate, We make a distribution adjustment for weights. Some neurons that do not satisfy this distribution are self autophagy. If the initialization parameter is uniform distribution, the weight are retained if the following conditions are satisfy with:
\begin{equation}
\widetilde{w_l}= \mathcal{F}_{dis} (w_l)
\end{equation}

\begin{equation}
w_l \in \lbrace x \sim \alpha \mathcal{U}(-\sqrt{ \frac{6}{n_l}},\sqrt{\frac{6}{n_l}}) \rbrace - \lbrace x \sim \mathcal{T}(\beta) \rbrace
\end{equation}

Where threshold $\alpha$ determine the range of values. When the weight is initialized as gaussian distribution, the condition as:
 \begin{equation}
w_l \in \lbrace x \sim \alpha \mathcal{N}(0,\frac{2}{n_l}) \rbrace - \lbrace x \sim \mathcal{T}(\beta) \rbrace
\end{equation}
After phagocytse neurons, those parameters is updated to $W_l \stackrel{update}{\longrightarrow}\widetilde{W_l}$, $n_l \stackrel{update}{\longrightarrow}\widetilde{n_l}$ and $h_l \stackrel{update}{\longrightarrow}\widetilde{h_l}$. Those parameters will used for next proliferate neorons, the autophagy condition would update as:

\begin{equation}
\begin{split}
Var(\widetilde{h_l} \otimes h_l^{'}) &= \mathit{Var}(\widetilde{h_1}\otimes h_1^{'}) \ \times \\
&(\prod\limits_{l = 1}^{L}\frac{1}{2}(\widetilde{n_l}+P_l) \times Var(\widetilde{W_l}\otimes W_l^{'}))
\end{split}
\end{equation}

\begin{equation}
\frac{1}{2}(\widetilde{n_l}+P_l)\mathit{Var}(\widetilde{W_l}\otimes W_l^{'})=1
\end{equation}

$h_l^{'}$, $w_l^{'}$ are new hidden layer activation and weighs value, respectivly. $\otimes$ is the concatenate, can operate in weights and activation values. $n_l=\widetilde{n_l}+\mathcal{P}_l$, $h_l \stackrel{growth}{\longrightarrow} \widetilde{h_l} \otimes h_l^{'} $, $W_l \stackrel{growth}{\longrightarrow} \widetilde{W_l} \otimes W_l^{'} $. From these definitions, With the metabolism of proliferation and autophagy, the weights, number of neurons and hidden layer activitions in the neurnal network would updated.

\textbf{Lower Representational Power of Neurons}
This part of the representational power of neurons is calculated in the same way as weights learning. In the weighs learning, we select the top $\mathcal{P}_l$ variance values from neurons in layer $l$. In this section, our model phagocytose neurons which its representational power below a certain level $ \nu$, the method is expressed as:
\begin{equation}
\widetilde{w_l}=\mathcal{F}_{low} (w_l,\nu)
\end{equation}

\textbf{Weight Density}
Density indicates the distribution of the data. According to the principle of fine density division, we subdivide the weights by ultra-fine density, referred to as $\mathcal{D}=\{D_1,D_2,\cdots, D_m\}$. The smaller the distance between weight samples, the more the density division area is. If the density of a certain area is much larger than other areas, the neurons in the area are randomly deleted the number $r$ of neurons. This method is a hypothesis and we will be verified in the experiment, it is expressed as:
 \begin{equation}
\widetilde{w_l}=\mathcal{F}_{den} (Max(\mathcal{D}),r)
\end{equation}

\section{Experiments}
In this sectoin, we present two basic architectures which are two fully-connected layers on MNIST, Fashion MNIST data sets and two-layers convolution neural network on CIFAR-10 to demonstrate the metabolism ability of our method, abbreviated as MetaNetNLP and MetaNetCNN. We discuss the experiment results, analyze the various parturition and autophagy behavior of our model on three different datasets. As for parameter calculation, we get average value of all epoch parameters with add new neurons.

\subsection{Results on MNIST}
The two hidden fully-connected network in this MNIST experiment have two basic architectures with BasicNLP(64) and BasicNLP(1024), that the number of neurons in all layers is 64 and 1024, respectively. Our model from the number of neurons in all layers is 32, 64 and 256 to constructive MetaNet, the final neuron number goal is no more than 1024 in each layers, those methods as MetaNetNLP(32-1024), MetaNetNLP(64-1024) and MetaNetNLP(256-1024).

\begin{figure}
\centerline{{\includegraphics[scale=0.25,clip,keepaspectratio]{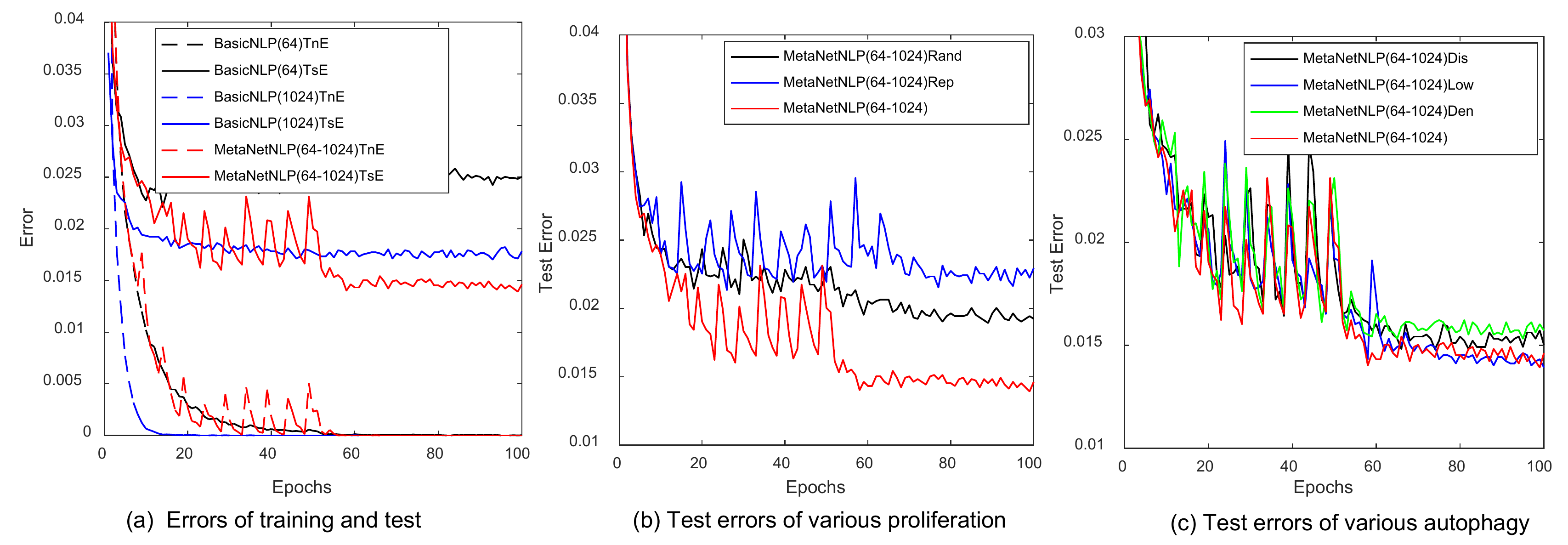}}}
\caption{Comparison of experimental results for basicline, neurons proliferation and autophagy in MNIST datasets. Left:(a) the dashed and solid line are the train and test error for different methods. Middle:(b) test error of various neurons proliferation methods. Right:(c) test error of diverse neurons autophagy methods.}
\label{fig_example3.ps}
\end{figure}

\begin{table}[!ht]
\footnotesize
\centering
\begin{tabular}{p{3.2cm}p{1.7cm}p{1cm}p{1.5cm}}%lllll
\toprule
Model & Test Error(\%) & Time(s) & \ \#Params(k) \\
\midrule
BasicNLP(64)   & 2.42 &135.87 & 54.26 \\
BasicNLP(1024) & 1.71 &185.27 & 1828.04 \\
\midrule
MetaNetNLP(32-1024) &1.48 &179.42 & 697.37\\
MetaNetNLP(64-1024) & \textbf{1.39} &180.46 & 735.76 \\
MetaNetNLP(256-1024) & 1.51 &184.63 & 895.91 \\
\bottomrule
\end{tabular}
\caption{Test error, training time and parameter on Mnist Datasets.}
\label{table1}
\end{table}

\ref{fig_example3.ps}(a) shows performance using BasicNLP(64), BasicNLP(1024) as the several different baseline for the neurons region, and MetaNetNLP(64-1024). The dashed line as train error(TnE),solid line as test error(TsE). During the training process, our method is slightly fluctuating due to the addition of neurons, but this has no effect on the final classification result. Generally, the performance of MetaNetNLP(64-1024) is not outstanding before 50 epochs, but at same point, such epoch 32,37 and 47, it is even better than BasicNLP(1024). After this critical point, the model becomes excellent, and the best results is 1.39\%, more than the 1.71\% of Basic1024. The final number of neurons in each layer of MetaNetNLP(32-1024), MetaNetNLP(64-1024) and MetaNetNLP(256-1024) is 942, 964 and 992, respectively. Combined with Table \ref{table1}, for the number of initialized neurons in the model, if we want to make the training time and resources cost less, that can set a smaller value, if want better results that can set a larger value. The fewer running time and neurons, parameter is less than half of BasicNLP(1024). Our method is better than baseline in terms of parameter, time and performance.

For MetaNet, the middle of Figure \ref{fig_example3.ps} we can see the role of different weight learning methods, the random parameterization as MetaNetNLP(64-1024)Rand, representational power of each neurons to all samples as MetaNetNLP(64-1024)Rep, the kernel ridge predictor we take it as a basic weights learning method, as MetaNetNLP(64-1024). From Figure \ref{fig_example3.ps}(b) kernel ridge predictor method is better. Figure \ref{fig_example3.ps}(c) is the different weights autophagy comparison method, weights selection based on distribution as MetaNetNLP(64-1024)Dis, the power of neurons representational as MetaNetNLP(64-1024)Low, the density as MetaNetNLP(64-1024)Den. These autophagy methods do not have a big impact on the experimental results. MetaNetNLP(64-1024) is self-growth but not ablate any neurous, relatively speaking, weights selection based on the power of neuronal representational is better than other methods for the MNIST dataset.

\begin{figure}
\centerline{{\includegraphics[scale=0.18,clip,keepaspectratio]{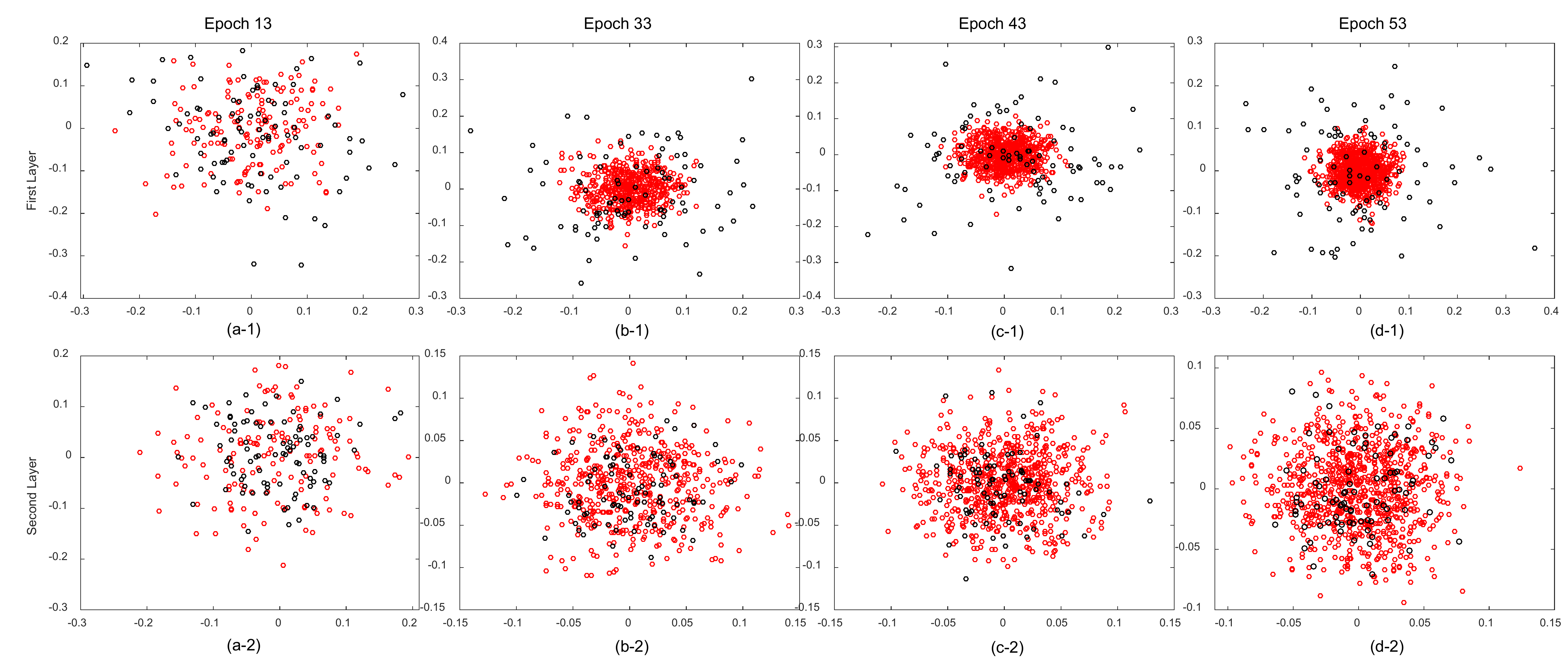}}}
\caption{Scatter plots of original and predicted parameters in different epochs and layers.}
\label{fig_example4.ps}
\end{figure}

During the metabolism of neurons, we use kernel ridge technology to learning the weights for proliferate neurons. Figure \ref{fig_example4.ps} shows the correlation between the original and predicted parameters in in different epochs and layers. We choose epoch 13,33,43 and 53 as a few samples in the training process to draw the dispersive points of corresponding weights. The upper half of Figure \ref{fig_example4.ps} are the first layer, at epoch 13 moment, the original and predicted parameters are more dispersed, then for epoch 33, 43 and 53, the original training parameters gathered together, but the predicted parameters are still dispersed, the weight diversity can learn as many features as possible on the initial layer, and provide more information for the next layer. For the bottom half of Figure \ref{fig_example4.ps}, this part are the parameters of second layer. Because our model has only two layers, the second layer learns the important feature representations for the task. Almost all predicted values are within the original parameter range, and gathered together. Compared with the predicted parameters in first layer, the prediction parameters of this layer are more conservative, maybe we can add some noise disturbance information in the parameter prediction process to improve the generalization of the model. But in any case, these prediction parameters can help MeteNet to training.

\subsection{Results on Fashion MNIST}
\begin{figure}
\centerline{{\includegraphics[scale=0.25,clip,keepaspectratio]{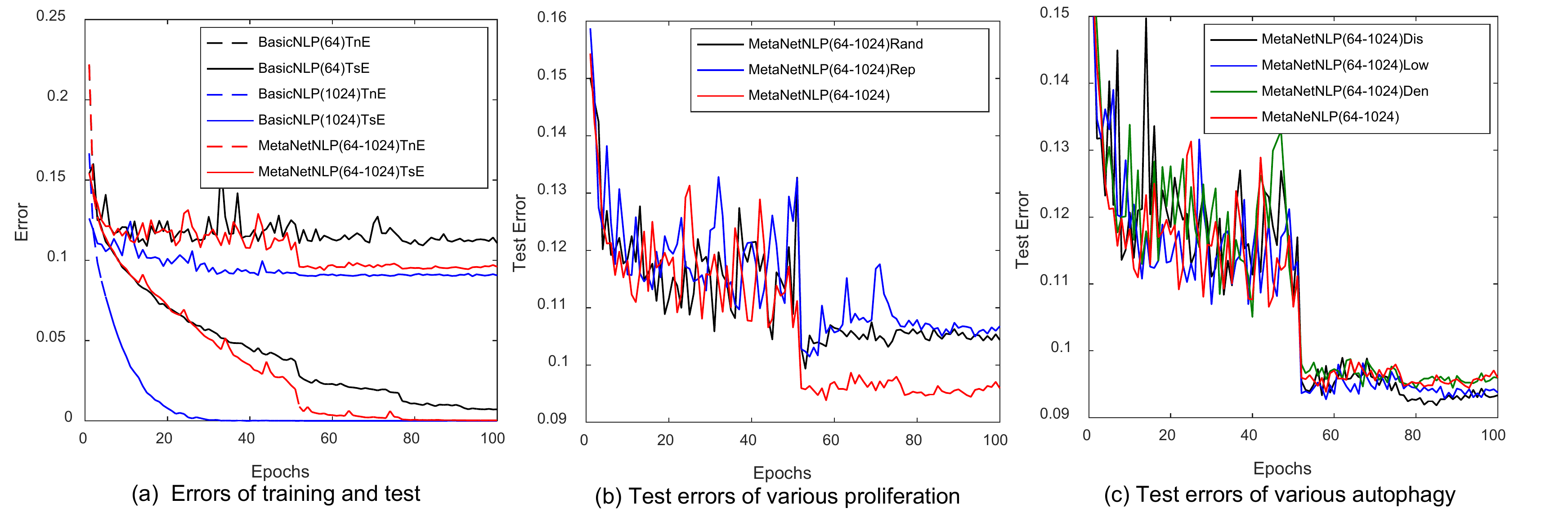}}}
\caption{Comparison of experimental results for basicline, neurons proliferation and autophagy in Fashion MNIST datasets. Left:(a) the dashed and solid line are the train and test error for different methods. Middle:(b) test error of various neurons proliferation methods. Right:(c) test error of diverse neurons autophagy methods.}
\label{fig_example5.ps}
\end{figure}

The basic model and parameter settings of Fashion MNIST are the same as MNIST. Figure \ref{fig_example5.ps}(a) shows the experimental results of BasicNLP(64), BasicNLP(1024) and MetaNetNLP(64-1024). For the different proliferate and ablate neurons methods, the experiments results Figure \ref{fig_example5.ps}(b) shows that random select some weights are better than the weights from higher representation power neurons. The higher representation power of the neurons may have insufficient ability to capture important features and have strong learning ability for less important features, which may result in poor performance. For the autophagy process, most methods are the same except that the method of ablating neurons based on distribution will converge faster and get the best results.

Table \ref{table2} shows the test error, cost time and parameters with different model. From the above results, the final test eror result BasicNLP(1024) is 9.06\%, MetaNetNLP(64-1024) is 9.34\%, MetaNetNLP(256-1024) is 9.02\%. For MetaNetNLP(64-1024) and MetaNetNLP(256-1024) only need fewer neurons and resources, can be almost achieve the same or even better results. It can be seen from the experimental results that our method save time and memory when achieve similar or even better results.

\begin{table}[!ht]
\footnotesize
\centering
\begin{tabular}{p{3.2cm}p{1.7cm}p{1cm}p{1.5cm}}%lllll
\toprule
Model & Test Error(\%) & Time(s) & \ \#Params(k) \\
\midrule
BasicNLP(64)   &10.93 &171.77 & 54.26 \\
BasicNLP(1024) & 9.06 &185.34 & 1828.04 \\
\midrule
MetaNetNLP(32-1024) & 9.44 & 174.18 & 697.37 \\
MetaNetNLP(64-1024) & 9.34 &177.58 & 735.76 \\
MetaNetNLP(256-1024) & \textbf{9.02} & 181.26 & 895.91   \\
\bottomrule
\end{tabular}
\caption{Test error, training time and parameter on Fashion Mnist Datasets.}
\label{table2}
\end{table}

\subsection{Results on CIFAR-10}
MetaNet on CIFAR-10 is different from the previous two datasets. We use two-layers convolution neural network to construct metabolism CNN. Due to the large amount of parameters in the CNN, the setting of smaller number of channels is adopted in the experimental model. The basic comparison method of the experiment is BasicCNN(60) with channel 60 and BasicCNN(80) with channel 80. Our model starts building itself from channel 40 and 60, as MetaNetCNN(40-80) and MetaNetCNN(60-80). The abbreviations of various experimental methods are the same as those on MNIST and Fashion MNIST. Our method finally increases the number of channels to 80.

\begin{figure}
\centerline{{\includegraphics[scale=0.25,clip,keepaspectratio]{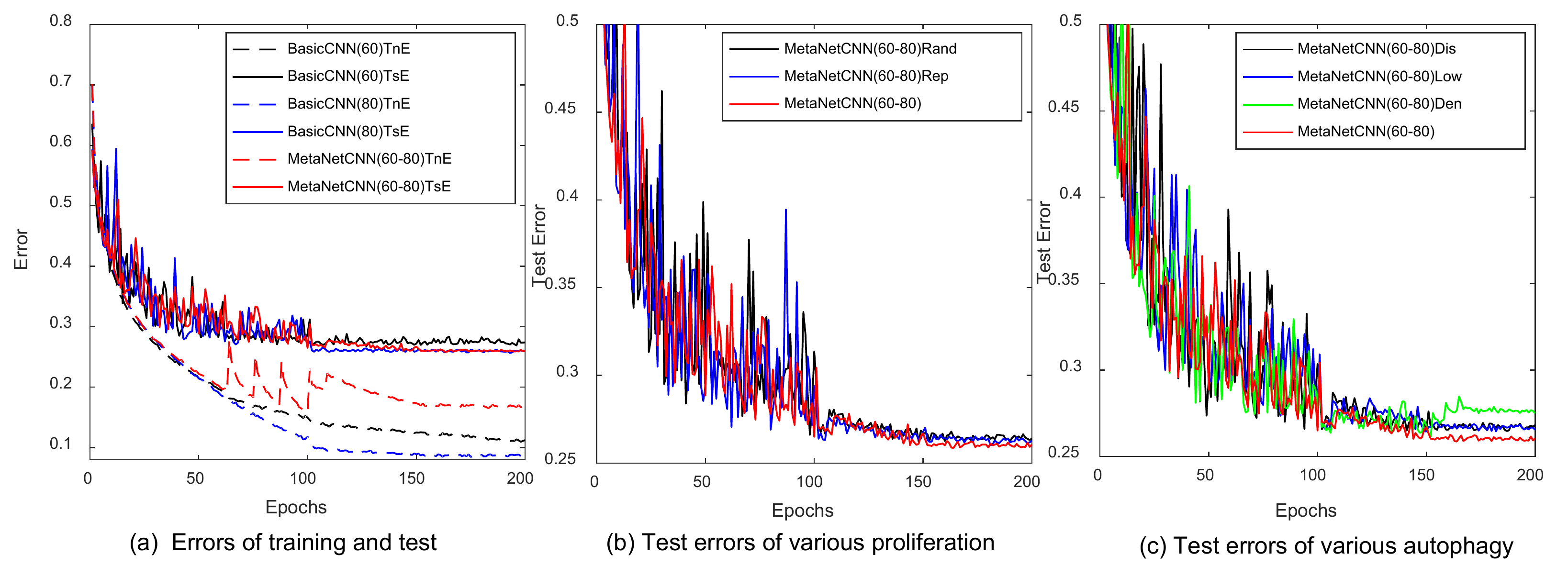}}}
\caption{Comparison of experimental results for basicline, neurons proliferation and autophagy in CIFAR-10 datasets. Left:(a) the dashed and solid line are the train and test error for different methods. Middle:(b) test error of various neurons proliferation methods. Right:(c) test error of diverse neurons autophagy methods.}
\label{fig_example6.ps}
\end{figure}

Table \ref{table3} shows the result of our MetaNetCNN(60-80) is 0.1 percentage points lower than BasicCNN(80), but in time it is 1200s faster than it, and the amount of parameters is also saved to about 200KB. Various operation on neruons proliferation and autophagy, we abbreviate the random parameterization as MetaNetCNN(60-80)Rand, representational power of each neurons to all samples as MetaNetCNN(60-80)Rep, the kernel ridge predictor we take it as a basic weights learning method, as MetaNetCNN(60-80), weights selection based on distribution as MetaNetCNN(60-80)Dis, the power of neurons representational as MetaNetCNN(60-80)Low, the density as MetaNetCNN(60-80)Den. Figure \ref{fig_example6.ps} mainly analysis the traing and test errors with basic and MetaNet methods, various methods of neurons proliferation and autophagy. Figure \ref{fig_example6.ps}(a) shows that MetaNet is relatively inferior in learning ability than BasicCNN(80), maybe more training iterations are needed in the process of neurons proliferation, but our model has good generalization ability. For the diverse proliferation methods, Figure \ref{fig_example6.ps}(b)(c) find that most of the methods are similar, but in the autophagy process, results in those three datasets show that the density of weights method is not good. Overall, our model has certain advantages that only need fewer resources when the required results are similar.

\begin{table}[!ht]
\footnotesize
\centering
\begin{tabular}{p{3.2cm}p{1.7cm}p{1cm}p{1.5cm}}%lllll
\toprule
Model & Test Error(\%) & Time(s) & \ \#Params(k) \\
\midrule
BasicCNN(60)  & 26.90 &3757.28 & 493.23 \\
BasicCNN(80) & \textbf{25.76} &4850.81 & 671.69 \\
\midrule
MetaNetCNN(40-80) & 26.70&3344.31 & 430.64 \\
MetaNetCNN(60-80) &25.86 & 3615.79& 499.80 \\
\bottomrule
\end{tabular}
\caption{Test error, training time and parameter on CIFAR-10 Datasets.}
\label{table3}
\end{table}

Through the above three experiments, we found the neurons proliferation of network can save a lot of time and parameter resources, for various parameters learning methods, the autoregression strategy have a good effect on weights prediction, the final effect on MNIST and Fashion MNIST datasets have exceed the BasicNLP(1024) method, and the number of neurons required and time is less. The different methods of neurons autophagy,
the experimental results based on density analysis are the worst, probably because some features of density aggregation are important, and random deletion has great uncontrollability, the important neurons are phagocytosed. Other based on the distribution of weights, the power of neuron representation experiments are similar, but ablating some neurons will save the time and parameters in the model.

\section{Conclusion}
We have introducted parturition and autophagy mechanism to metabolize neural network that automatically self growth and ablate neurons in each layers. The idea of MetaNet are verify on two fully-connected layers and two-layers convolution neural network, the experiments have demonstrated the benefits of our model. For deliver dynamic neurons number, we intend to study better function expression where each layers can get the more adaptable neurons number. The autophagy in our model didn't play a better role, we plan to design a appropriate autophagy mechanism that reflect the effectiveness of its behavior. The framework can applicable to different architectures and generalizability is well in theory, furthermore, we could extend it multiple layers network and test its generalization by more data sets.

\newpage

%\subsection*{References}
\bibliographystyle{AAAI}
\bibliography{bibfile}

\end{document}